\documentclass[runningheads]{llncs}
\usepackage{graphicx}
\usepackage{main}
\usepackage{amsmath,amssymb} % define this before the line numbering.
\usepackage{color}

\usepackage{adjustbox}
\usepackage{booktabs} 
\usepackage{threeparttable}
\usepackage[table]{xcolor}
\usepackage{threeparttable}
\usepackage{multirow}
\usepackage{enumitem} 
\usepackage{xcolor}
\usepackage{tabularx}
\usepackage{hyperref}

\begin{document}
% \renewcommand\thelinenumber{\color[rgb]{0.2,0.5,0.8}\normalfont\sffamily\scriptsize\arabic{linenumber}\color[rgb]{0,0,0}}
% \renewcommand\makeLineNumber {\hss\thelinenumber\ \hspace{6mm} \rlap{\hskip\textwidth\ \hspace{6.5mm}\thelinenumber}}
% \linenumbers
\pagestyle{headings}
\mainmatter
\def\ECCVSubNumber{3215}  % Insert your submission number here

\title{iCaps: An Interpretable Classifier \\
via Disentangled Capsule Networks} % Replace with your title

% INITIAL SUBMISSION 
%\begin{comment}
%\titlerunning{ECCV-20 submission ID \ECCVSubNumber} 
%\authorrunning{ECCV-20 submission ID \ECCVSubNumber} 
%\author{Anonymous ECCV submission}
%\institute{Paper ID \ECCVSubNumber}
%\end{comment}

% CAMERA READY SUBMISSION
%\begin{comment}
\titlerunning{iCaps: An Interpretable Classifier via Disentangled Capsule Networks}
% If the paper title is too long for the running head, you can set
% an abbreviated paper title here
% 0000-0003-4521-2958
% \orcidID{0000-0002-1344-1054}
% \orcidID{0000-0002-1530-1020}
% \orcidID{0000-0001-5762-6643}
% \orcidID{0000-0003-4521-2958}
\author{Dahuin Jung \and
Jonghyun Lee \and Jihun Yi \and
Sungroh Yoon\thanks{Correspondence to: Sungroh Yoon}}

\authorrunning{D. Jung et al.}
% First names are abbreviated in the running head.
% If there are more than two authors, 'et al.' is used.
%
\institute{Electrical and Computer Engineering, \\ ASRI, INMC, and Institute of Engineering Research \\ Seoul National University, Seoul 08826, South Korea \\
\email{\{annajung0625, leejh9611, t080205, sryoon\}@snu.ac.kr}}
%\end{comment}

\maketitle

\begin{abstract}
We propose an interpretable Capsule Network, $\textit{iCaps}$, for image classification. A capsule is a group of neurons nested inside each layer, and the one in the last layer is called a class capsule, which is a vector whose norm indicates a predicted probability for the class. Using the class capsule, existing Capsule Networks already provide some level of interpretability. However, there are two limitations which degrade its interpretability: 1) the class capsule also includes classification-irrelevant information, and 2) entities represented by the class capsule overlap. In this work, we address these two limitations using a novel class-supervised disentanglement algorithm and an additional regularizer, respectively. Through quantitative and qualitative evaluations on three datasets, we demonstrate that the resulting classifier, $\textit{iCaps}$, provides a prediction along with clear rationales behind it with no performance degradation.

\end{abstract}

\section{Introduction}
Despite the success of deep learning in a broad range of tasks, including image classification and segmentation, speech synthesis, and medical decision-making, the reliability of decisions made by artificial intelligence is still questionable. Hence, many promising studies have been conducted regarding explainable artificial intelligence (XAI). The main task of XAI is to provide explanations that can aid the comprehension of provided decisions to users. Using these explanations, users can check whether a model performs as expected or identify potential bias/problems inherent to the model.

Several different approaches have been proposed to explain deep learning models. In some studies, models that can provide human-understandable explanations of their predictions without retraining or modification have been proposed. These studies aim for built-in interpretability. We herein propose a new built-in interpretable model that offers a concept-based explanation using Capsule Networks (CapsNets)~\cite{capsulenet}. 

As the main building block, CapsNets use capsules - a group of neurons – that encapsulates the instantiation parameters of an entity, such as an object or its fragments. The magnitude of the output vector of a capsule indicates the probability that the encoded instantiation parameter is present in the input. The capsules in the final layer are called class capsules, and the norm of each class capsule indicates the predicted probability of each class. The instantiation parameters of the class capsule can represent the position, color, texture, and scaling of an object or its fragments, and these can be interpreted as concepts to humans.

Therefore, the instantiation parameters represented by the class capsule and its magnitudes can be used, to an extent, to explain a model’s prediction. However, two factors degrade interpretability. First, some instantiation parameters of the class capsule represent classification-irrelevant concepts. Next, a single concept can be encoded in multiple elements of the class capsule. Therefore, a single concept can be represented by different magnitudes in two different elements. 

\begin{figure*}[t]
\centering
\includegraphics[width=0.9\linewidth]{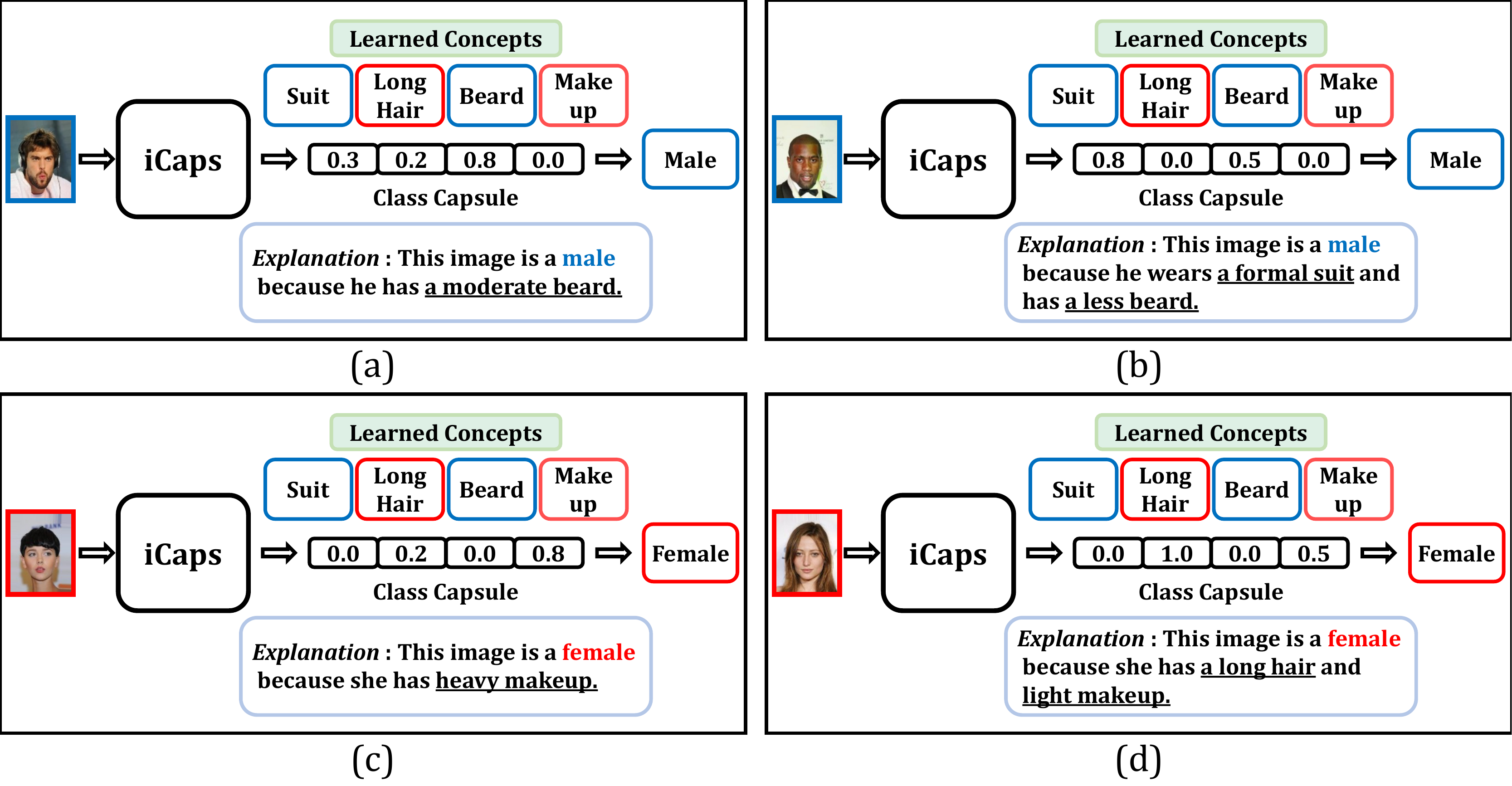}
\caption{Overview of our study. We propose a new interpretable classifier, iCaps, which classifies an observation by only considering class-relevant variables; these class-relevant variables are human-understandable concepts. By analyzing the values of the class-relevant variables (concepts), we can understand the decisions made by iCaps.}
\label{fig:goal}
\end{figure*}

By addressing these two problems, we propose an interpretable CapsNet architecture, $\textit{iCaps}$, that only contains classification-relevant distinct concepts in the class capsule. The overview of iCaps is described in Fig.~\ref{fig:goal}. To address the first problem, we propose a novel class-supervised disentanglement method that disentangles class-relevant and -irrelevant features within an observation, without any leakage. For the second problem, we use an additional regularizer based on latent traversal to prevent the same concept from being encapsulated several times in the class capsule.

Some built-in interpretable models use predefined concepts to provide explanations~\cite{tcav,tcav2}. However, such prior knowledge is not available or costly to define in most cases; hence, in this study, we assumed where the concepts were learned instead. Moreover, we posit three desiderata for an interpretable classifier based on the learned concepts: informativeness, distinctness, and explainability where, for example, informativeness ensures that only classification-relevant information is used to provide an explanation of the model's prediction. Based on the three desiderata, we validated our model both theoretically and empirically. Our main contributions in this study are as follows:
\begin{enumerate}
  \item We improve the explainability property of CapsNets by addressing two problems: classification-irrelevant information and overlapping.
    \item We suggest a novel class-disentanglement algorithm that can disentangle the latent feature of $x$ into two complementary subspaces: class-relevant and -irrelevant subspaces. The class-relevant subspace of our algorithm contains intra-class variation, unlike prior studies, which contain only inter-class variation.
    \item We posit three desiderata for an interpretable classifier based on learned concepts and demonstrate the effectiveness of iCaps based on the three desiderata.
\end{enumerate}

\section{Related Work}
\subsection{Capsule Networks}
\label{section:capsulenetworks}
CapsNet~\cite{capsulenet} is a neural network based on a group of neurons – a vector. The original CapsNet has a simple network structure comprising three layers. First, an observation $x$ undergoes a convolution layer to transfer pixel-level information into a latent space, followed by the Primary-Capsule (PC) layer and Class-Capsule (CC) layer. Information contained in the PC layer is transferred to the CC layer above using a dynamic routing method, and this method is called “routing by agreement". The coupling coefficients between the capsules in the PC and CC layers are updated in a direction that can increase the classification performance (a top-down mechanism). The output of the CC layer is a class capsule of classes, and the norm of each class capsule indicates the predicted probability for each class. The elements of the class capsules represent the instantiation parameters of a type of entity. To encode these instantiation parameters for the class capsules, margin loss and reconstruction loss are used. The margin loss, $\mathcal{L}_M$, is:
\begin{equation}
    \begin{aligned}
      \mathcal{L}_{M} = \lambda_{M} \: (y^{i} \: \text{max} (0,\, m^+ - \left \| c \right \|)^2 + 0.5 \: ((1 - y^i) \: \text{max} (0,\,  \left \| c^{\neq i} \right \| - m^-)^2)),
   \label{eq:margin_loss}
   \end{aligned}
\end{equation}
where $y^i$ = 1 iff the ground-truth class label is $i$, $c$ is the class capsule for the ground-truth class $i$, $c^{\neq i}$ are the class capsules except for $c$, $m^+$ = 0.9, and $m^-$ = 0.1. The reconstruction loss, $\mathcal{L}_{\text{recon}}$, is used, which is expressed as
\begin{equation}
    \begin{aligned}
      \mathcal{L}_{\text{recon}} = \lambda_{\text{recon}} \: \mathbb{E}\left \| \hat{x} - x \right \|_{F}^{2},
   \label{eq:recon_loss}
   \end{aligned}
\end{equation}
where $\hat{x}$ is reconstructed $x$ using an additional decoder, which uses $c$ as the input. The original CapsNet presents some computational and structural limitations. Hence, some advanced studies based on CapsNet have been performed~\cite{starcaps,selfroutcapsule,laddercapsule,deepcaps}. Among these, our study uses DeepCaps~\cite{deepcaps} as a base because it yields better classification performance than the original CapsNet on more complex images by utilizing a skip connection and a three-dimensional convolution-inspired routing method and is easy to apply to our work. More detailed information regarding CapsNets is available in~\cite{capsulenet,deepcaps}.

\subsection{Disentanglement} 
Our work utilizes class-supervised disentanglement to create an interpretable classifier that provides an explanation using only classification-relevant information. Class-supervised disentanglement learning aims to disentangle the latent feature of $x$ into two complementary subspaces - class-relevant and -irrelevant subspaces - in a setting where the class label for images in the training set is provided. Two approaches can be used in class-supervised disentanglement: adversarial and non-adversarial.
DrNet~\cite{drnet} and Szabo et al.~\cite{szabo} are adversarial methods, whereas Cycle-VAE~\cite{cyclevae}, ML-VAE~\cite{mlvae}, and LORD~\cite{lord} are non-adversarial methods. Implicitly or explicitly, all class-supervised disentanglement methods assume that inter-class variation is much larger than intra-class variation; therefore, intra-class variation can be ignored. On the contrary, our work assumes that intra-class variation should not be ignored even though it is relatively small. From the perspective of an interpretable classifier, intra-class variation is an important feature to explain a model's prediction. Also, our method has some level of similarity with InfoGAN~\cite{infogan} (unsupervised disentanglement method) in a structural way. The comparison is given in Sec. S10 of the supplementary.

\subsection{Interpretable Methods}
Two topics of research in XAI provide two different notions of interpretability of deep models: (1) post-hoc interpretability and (2) built-in interpretability.
(1) Post-hoc interpretability methods aim to interpret models or decisions of already trained neural networks. By contrast, networks that are inherently interpretable provide (2) built-in interpretability.

\subsubsection{Post-hoc Interpretability}
Starting from the Saliency map~\cite{saliencymap}, a number of post-hoc interpretation methods have been suggested to visually explain the decision of a classifier. These methods generate a heatmap of the same size as the input image and highlight the decisive regions within the input image. Post-hoc methods are based on backpropagation~\cite{vargrad,lrp,gradcam,deeplift,saliencymap,smoothgrad,gbp,integratedgradients}, local perturbation~\cite{occlusion,pda}, or mask optimization~\cite{counterfactual,realtime,extremalpert,meaningfulpert,rise}.
The resulting heatmap of post-hoc methods are easily interpretable. However, evaluating the quality of their results is non-trivial, and both the method and the evaluation metrics are active research areas. Recently, several interpretable methods that offer explanations based on concepts and prototypes have been proposed~\cite{tcav,senn,protopv1,protopv2}. TCAV~\cite{tcav} is a post-hoc method based on predefined concepts. TCAV offers an explanation by finding the closest predefined concepts to the corresponding class in the feature space. Unlike most post-hoc methods, TCAV only offers an explanation for a class, not for a single data point.

\subsubsection{Built-in Interpretability} 
SENN~\cite{senn} suggests an interpretable classifier structure that predicts a class by combining concepts and relevances. SENN encodes input features into two representations: concepts and relevances. The importance of a given concept for classification can be explained through the relevance score of the corresponding concept. However, the learned concepts of SENN are not clearly human-understandable, as analyzed in Sec. S3 of the supplementary. ProtoPNet~\cite{protopv2} proposes an interpretable classifier based on prototypes. ProtoPNet learns prototypical patches of each class from the training dataset. After finding the prototypes, the model makes a decision by measuring the distance between local patches of the test observation and the found prototypes of each class.

\section{iCaps: An Interpretable Classifier via Disentangled Capsule Networks}
\begin{figure*}[t]
\centering
\includegraphics[width=0.9\linewidth]{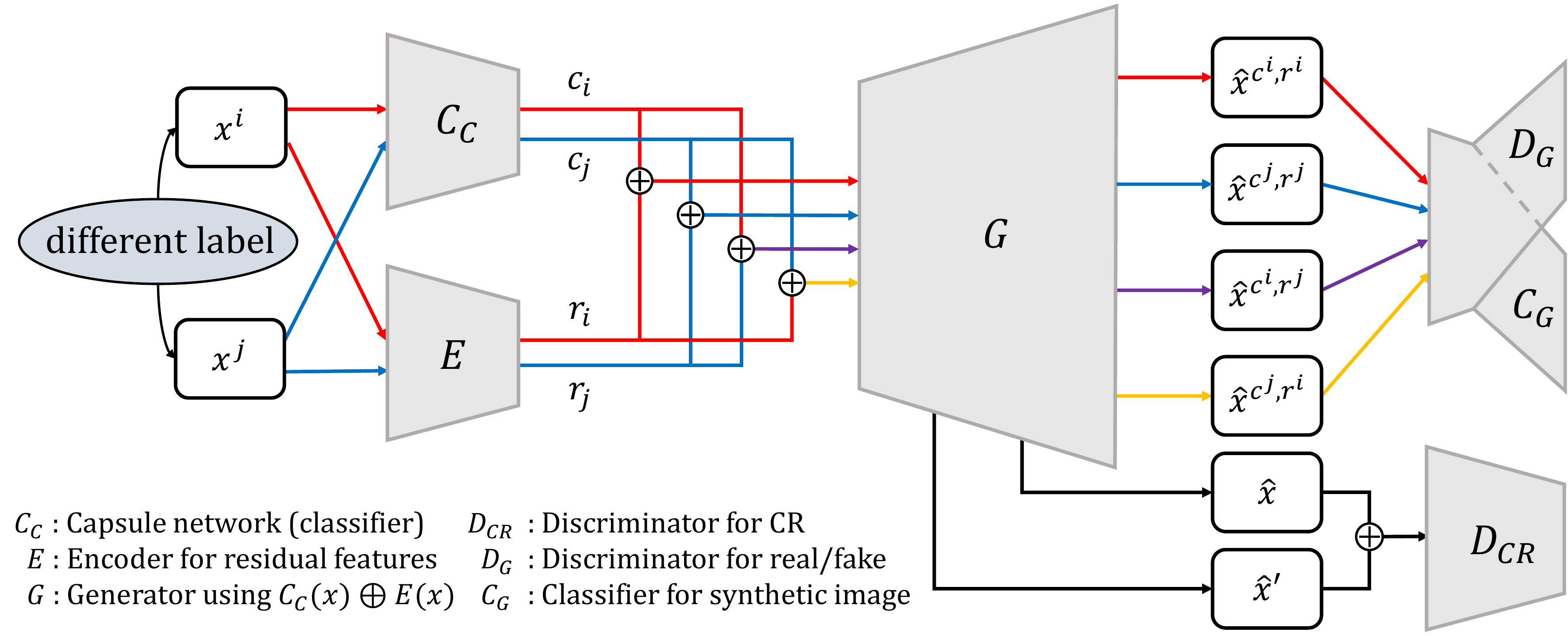}
\caption{Network architecture. $x^i$ and $x^j$ are two images of different labels. $C_C$ encodes only class-relevant features of $x^i$ and $x^j$ to $c^i$ and $c^j$, respectively. $E$ encodes only class-irrelevant (residual) features of $x^i$ and $x^j$ to $r^i$ and $r^j$, respectively. $G$ constructs the images $\hat{x}^{c^i, r^i}$ and $\hat{x}^{c^j, r^j}$ using ${c^i \oplus r^i}$ and ${c^j \oplus r^j}$, respectively. Also, $G$ generates the images $\hat{x}^{c^i, r^j}$ and $\hat{x}^{c^j, r^i}$ using swapped ${c^i \oplus r^j}$ and ${c^i \oplus r^j}$, respectively. These images are distinguished as real or fake and classified by $D_G$ and $C_G$. $D_{\text{CR}}$ takes two images $\hat{x}$ and $\hat{x}'$ that share the same $c$ and $r$ except for a single index $l$ of $c$, and it is trained to identify $l$.}
\label{fig:icaps}
\end{figure*}

iCaps comprises six components, as illustrated in Fig.~\ref{fig:icaps}.

\begin{itemize}[leftmargin=5.5mm, labelindent=7.5mm,labelsep=3.3mm]
\item $C_C$: a capsule network (classifier) that represents the class-relevant latent space.

\item $E$: an encoder that represents the class-irrelevant (residual) latent space.

\item $G$: a generator that creates synthetic images using $C_C$($x$) $\oplus$ $E$($x$), where $\oplus$ represents concatenation.

\item $D_G$: a discriminator for image generation, that distinguishes whether an observation is from the dataset or from $G$.

\item $C_G$: a classifier for image generation, that estimates class labels.

\item $D_{\text{CR}}$: a discriminator for contrastive regularization (CR), that maximizes the distance between the concepts represented by $C_C$

\end{itemize}
Assume that a collection of $n$ images $x_1, x_2, ..., x_n$ $\in \mathcal{X}$ and their corresponding labels $y_i$ $\in \left [ k \right ]$ is provided. $k$ and [$k$] (=[1,…,$k$]) are the number and the set of classes, respectively. $\mathcal{X}^{i}$ represents all images corresponding to a class index $i$. As described in Fig.~\ref{fig:icaps}, iCaps uses two images of different class labels as input in the training phase. In the case of binary classification, the input pairs are images of the opposite class labels. In multiclass classification, two class labels are randomly selected in each batch.

We assume that the representation of images can be disentangled into two complementary latent spaces, $\mathcal{C} $ and $\mathcal{R}$. Our objective is to find a class-relevant representation $c_i$ $\in \mathcal{C}$ and a class-irrelevant (we call as residual) representation $r_i$ $\in \mathcal{R}$ for each image $x_i$. The size of output vector of the class-relevant representation is $L$.

\subsection{Disentanglement between Class-relevant and Class-irrelevant Information}
\label{section:disentanglement}
The class-relevant subspace, $\mathcal{C}$, is represented by $C_C$, and the residual subspace, $\mathcal{R}$, is represented by $E$. The class-relevant representation, $c_i$, contains all the information relevant for classification, whereas the residual representation, $r_i$, contains residual information irrelevant for classification. The previous class-supervised disentanglement methods~\cite{drnet,cyclevae,mlvae,lord} contain only information shared by each class (inter-class variation) in $c_i$, assuming that $\left \| c_i - c_j \right \|_{F}^{2} = 0$ if $y_i = y_j$. However, under this assumption, $c_i$ cannot include classification-relevant intra-class variation. For example, if a model is trained using a dataset labeled as female and male, and most of the males are wearing suits, then wearing a suit should be a classification-relevant variable. In other words, this variable should be included in $c$. However, to satisfy the assumption: $\left \| c_i - c_j \right \|_{F}^{2} = 0$ if $y_i = y_j$ above, every man should be defined as wearing a suit. This is not true. If this variable is not included in $c$, it should be included in $r$. Consequently, information leakage is implied because wearing a suit is classification-relevant information in this model.

We wish to include classification-relevant intra-class variation in $c_i$. By analyzing only $c_i$, we can understand the rationale behind the model’s prediction. More information-theoretically, when the mutual information between $c$ and $y$ is:
\begin{equation}
    \begin{aligned}
      I(c;\: y) = \int_y \int_c p(y) \, q(c|y) \, \text{log} \, \frac{q(c|y)}{q(c)} \, dc\,dy,
   \label{eq:ml_c}
   \end{aligned}
\end{equation}
where $q(c) = \int_y p(y) \, q(c|y) \, dy$, $I(c;\: y)$ should be non-zero, and $I(r;\: y)$ should be zero. To obtain $I(c;\: y) > 0$ and $I(r;\: y) = 0$, we utilize three objective functions. 

The first objective is a cross-entropy loss that includes two images generated by swapping $r$ of two inputs. The loss term is:
\begin{equation}
    \begin{aligned}
    \mathcal{L}_{C_G} = \lambda_{C_G} \: (\mathbb{E}& [ (1 - y_t)\cdot  C_G(x)] + \mathbb{E} [ (1 - y_{t}^{i})\cdot C_G({\hat{x}^{c^i,r^i}})] + \\ & \mathbb{E} [ (1 - y_{t}^{i})\cdot C_G({\hat{x}^{c^i,r^j}})]  -
    \mathbb{E} [y_{t} \cdot C_G(x)]] - \\ & \mathbb{E} [y_{t}^{i} \cdot C_G(\hat{x}^{c^i,r^i})]] - \mathbb{E} [y_{t}^{i} \cdot C_G({\hat{x}^{c^i,r^j}})]]),
    \label{eq:cls_loss}
   \end{aligned}
\end{equation}
where $i$ and $j$ are two different class indices, $x\sim p_{\text{data}}(x)$, $x^i\sim p_{\text{data}}(x^i)$, $x^j\sim p_{\text{data}}(x^j)$, $y_t$ is an one-hot encoding of $y \in[k]$, $\hat{x}^{c^i,r^i}\sim G(C_C(x^{i}), E(x^{i}))$, and $\hat{x}^{c^i,r^j}\sim G(C_C(x^{i}), E(x^{j}))$. The class labels of the two synthetic images, $\hat{x}^{c^i,r^i}$ and $\hat{x}^{c^i,r^j}$, are the same as $y_{t}^{i}$. For all images in Eq. 4, the higher the probability of class $i$, the smaller the loss.

Furthermore, we use the prediction confidence of the model as a loss. We propose a class-similarity (CS) loss: 
\begin{equation}
    \begin{aligned}
    \mathcal{L}_{\text{CS}} = \lambda_{\text{CS}} \: \mathbb{E}  \left \| C_{G}^{\text{logit}}(\hat{x}^{c^i,r^i}) - C_{G}^{\text{logit}}(\hat{x}^{c^i,r^j}) \right \|_{F}^{2},
   \label{eq:class_sim}
   \end{aligned}
\end{equation}
where $C_{G}^{\text{logit}}$ represents the logit of $C_{G}$. In Eq. 5, we can compare the likelihood distributions of two images generated based on the same $c$ yet different $r$. That is, the images generated with the same $c$ should have exactly the same likelihood distributions in our framework. $\mathcal{L}_{C_G}$ and $\mathcal{L}_{\text{CS}}$ cause information relevant to the classification to be included in $c$.

The third loss is for the residual features independent of classification. Other similar studies use either an adversarial loss~\cite{drnet}, KL-divergence term~\cite{mlvae}, or asymmetric noise regularization~\cite{lord} to encode class-irrelevant features in $r$. Unlike these methods, we allow $r$ to include the remaining information by causing $c$ to include all the relevant information for classification.  The residual similarity (RS) loss is defined as follows:
\begin{equation}
    \begin{aligned}
    \mathcal{L}_{\text{RS}} = \lambda_{\text{RS}} \: \mathbb{E} \left \| E(x^{i}) - E(\hat{x}^{c^j,r^i}) \right \|_{F}^{2}.
   \label{eq:residual_sim}
   \end{aligned}
\end{equation}
To minimize $\mathcal{L}_{\text{RS}}$, the class-irrelevant (residual) feature, $r$, of the real image $x^{i}$ corresponding to class index $i$ and the class-irrelevant (residual) feature, $r$, of $\hat{x}^{c^j,r^i}$ should be the same. This is only possible when $c$ contains all the classification-relevant information.

In addition, in our study, we assume that all dimensions of $c$ are used to represent class-relevant concepts. More formally, 
\begin{equation}
    \begin{aligned}
    \underset{l \: \in \: L}{\text{min}} \: \mathbb{E} [I(C_C(x); y)_l] > 0
    \label{eq:all_c}
   \end{aligned}
\end{equation}
where $L$ is the size of $c$. However, $\mathcal{L}_{M}$, $\mathcal{L}_{C_G}$ and $\mathcal{L}_{\text{CS}}$ do not guarantee the above assumption. Therefore, we propose the loss as follows:
\begin{equation}
    \begin{aligned}
    \mathcal{L}_{\text{concept}} = -\: \lambda_{\text{concept}} \: ( \underset{l \: \in \: L}{\text{min}} \: \left |  \mathbb{E} [C_C(x^{i})] - \mathbb{E} [C_C(x^{j})] \right |_l).
    \label{eq:concept_loss}
   \end{aligned}
\end{equation}
$\mathcal{L}_{\text{concept}}$ is additionally suggested to maximize the mutual information between all elements of $c$ and $y$. By using $\mathcal{L}_M$ and $\mathcal{L}_{\text{concept}}$ together, we enforce all dimensions of $c$ to represent discriminative concepts by maximizing the smallest distance between different classes $i$, $j$ among all the dimensions. We elaborate the benefits of the proposed losses and components in Secs.~\ref{section:experiments} and S2.

\subsection{Latent Traversal}
\label{section:latent_traversals}
We add a discriminator, $D_{\text{CR}}$, to prevent overlapping concepts, which is the second problem we wish to solve. The idea of $D_{\text{CR}}$ comes from latent traversal. Latent traversal refers to generating images by traversing a single element of a latent space; it is widely used to measure disentanglement in evaluation~\cite{betavae,factorvae}. Lin et al.~\cite{infogancr} first used the idea of latent traversal in the training phase, and named as contrastive regularization. By following it, we also call our loss as $\mathcal{L}_{\text{CR}}$. $\mathcal{L}_{\text{CR}}$ can be expressed as follows:
\begin{equation}
    \begin{aligned}
\mathcal{L}_{\text{CR}} = - \lambda_{\text{CR}} \: \, \mathbb{E}_{l\sim U[L],\:  (\hat{x}, \hat{x}') \sim G(C_C(x), E(x))} \left [ \left \langle I, \: \text{log} \, D_{\text{CR}} (\hat{x}, \hat{x}') \right \rangle \right ],
    \label{eq:cr_loss}
   \end{aligned}
\end{equation}
where $l$ is a random index over $L$, $\hat{x}$ and $\hat{x}'$ are two images generated with different value of $C_C{(x)}_l$ while fixing the remaining elements. $<\cdot>$ represents a dot product, and $I$ denotes the one-hot encoding of the random index $l$. $\mathcal{L}_{\text{CR}}$ forces changes in the elements of $c$ to be visually noticeable and easy to distinguish between each other. The difference between our $L_{\text{CR}}$ and that reported in \cite{infogancr} is that we directly used the definition of latent traversal. However, Lin et al.~\cite{infogancr} fixed a single element and changes all the remaining elements, and tries to identify the fixed element.

\subsection{Interpretable Classifier Based on Learned Concepts}
\label{section:interpretable_classifier}
In addition, we replace the decoder part of DeepCaps with Generative Adversarial Networks (GANs)~\cite{gan} to encourage $\hat{x}^{c^i,r^j}$ to be realistic. To enable $\mathcal{L}_{C_G}$ and 
$\mathcal{L}_{\text{CS}}$ to function as intended, the quality of generated image $\hat{x}^{c^i,r^j}$ is important. Unlike $\hat{x}^{c^i,r^i}$, $\hat{x}^{c^i,r^j}$ does not have a ground-truth image. Therefore, we used an adversarial game of GANs and the ACGAN~\cite{acgan} structure; as such, $C_G$ and $D_G$ share the weight except for the last fully connected layer. The losses for $G$ and $D_G$ are as follows:
\begin{equation}
    \begin{aligned}
    \mathcal{L}_{G} = - \lambda_{G} \: \mathbb{E} [D_{G}(\hat{x})], \;\; \mathcal{L}_{D_{G}} = \lambda_{D_G} \: (\mathbb{E} [D_G{(\hat{x})}] - \mathbb{E} [D_G{(x)}]).
    \label{eq:generator_loss}
    \end{aligned}
\end{equation}
$L_G$ is a loss to create a realistic image, and $L_{D_G}$ is a WGAN~\cite{wgan} based loss to distinguish generated images by $G$ from real images. For the gradient penalty, we used a Lipschitz gradient penalty term~\cite{lgp}:
\begin{equation}
    \begin{aligned}
    \mathcal{L}_{\text{LGP}} = \lambda_{\text{LGP}} \:\mathbb{E} (\left \| \bigtriangledown_{\hat{x}} D_{G}(\hat{x}) \right \|_2 - 1)_{+}^{2}.
    \label{eq:gp_loss}
    \end{aligned}
\end{equation}

The built-in interpretable model domain is still new. By analyzing similar studies, we posit a reasonable set of desiderata for an interpretable classifier based on learned concepts as follows:
\begin{enumerate}
    \item \textbf{Informativeness}: the concept representation of $x$ for explanations should preserve only classification-relevant information,
    \item \textbf{Distinctness}: the learned concepts should be non-overlapping,
    \item \textbf{Explainability}: a decision should be explained with human-understandable concepts.
\end{enumerate}
We obtained these conditions by (i) encoding only the class-relevant information in $c$, (ii) enforcing distinctness by an additional discriminator and (iii) exploiting the fact that the instantiation parameters represented by the class capsule are human-understandable concepts.

\begin{table}[t]
\centering
\caption{Classification accuracy ($C_C$) of $p(y|c)$ (higher is better).}
\begin{tabular}{@{}ccccc@{}}
\toprule
Architecture         & MNIST  & SVHN   & CelebA &  \\ \midrule
ResNet-18~\cite{resnet}  & 0.992 & 0.945 & 0.977 &  \\
DeepCaps~\cite{deepcaps}  & 0.997 & 0.971 & 0.974 &  \\
Ours      & 0.992  & 0.920   &  0.984  &  \\ \bottomrule
\label{acc}
\end{tabular}
\end{table}

To train iCaps, we alternatively trained $C_C$, $E$, $D_G$ $\&$ $C_G$, $G$, and $D_{\text{CR}}$ using the following gradients:
\begin{equation}
    \begin{aligned}
    \theta_{C_C} \overset{+}{\leftarrow} -\Delta_{\theta_{C_C}} ( \mathcal{L}_{M} +  \mathcal{L}_{\text{recon}} +  \mathcal{L}_{\text{concept}} +   \mathcal{L}_{C_G} +  \mathcal{L}_{\text{CS}} +  \mathcal{L}_{\text{RS}} + \mathcal{L}_{\text{CR}})
    \label{eq:gradient_l_c_c}
    \end{aligned}
\end{equation}
\begin{equation}
    \begin{aligned}
    \theta_{E} \overset{+}{\leftarrow} -\Delta_{\theta_{E}} ( \mathcal{L}_{\text{recon}} +  \mathcal{L}_{\text{KL}} + \mathcal{L}_{\text{CS}} + \mathcal{L}_{\text{RS}} + \mathcal{L}_{C_G})
    \label{eq:gradient_l_e}
    \end{aligned}
\end{equation}
\begin{equation}
    \begin{aligned}
    \theta_{(D_G, C_G)} \overset{+}{\leftarrow} -\Delta_{\theta_{(D_G, C_G)}} (\mathcal{L}_{D_G} + \mathcal{L}_{C_G} + \mathcal{L}_{\text{CS}} + \mathcal{L}_{\text{LGP}})
    \label{eq:gradient_l_dg_cg}
    \end{aligned}
\end{equation}
\begin{equation}
    \begin{aligned}
    \theta_{G} \overset{+}{\leftarrow} -\Delta_{\theta_G} ( \mathcal{L}_{G} + \mathcal{L}_{C_G} +  \mathcal{L}_{\text{recon}} + \mathcal{L}_{\text{CS}} +  \mathcal{L}_{\text{RS}} + \mathcal{L}_{\text{CR}})
    \label{eq:gradient_l_g}
    \end{aligned}
\end{equation}
\begin{equation}
    \begin{aligned}
    \theta_{D_{\text{CR}}} \overset{+}{\leftarrow} -\Delta_{\theta_{D_{\text{CR}}}} \mathcal{L}_{\text{CR}}
    \label{eq:gradient_l_cr}
    \end{aligned}
\end{equation}
$\mathcal{L}_{\text{KL}}$ represents the KL term of a variational autoencoder~\cite{vae}; $\mathcal{L}_{\text{KL}}$ is scaled down by a small hyperparameter such that it does not reduce the reconstruction ability~\cite{factorvae}. $\mathcal{L}_{M}$ and $\mathcal{L}_{\text{recon}}$ are provided in Sec.~\ref{section:capsulenetworks}. In case of $\mathcal{L}_{\text{recon}}$, $\hat{x}$ is from $G(C_C(x), E(x))$.

\section{Experiments}
\label{section:experiments}
We evaluated the performance of our method and the comparison methods on three datasets: MNIST~\cite{mnist} (digit number as a class label), SVHN~\cite{svhn} (digit number as a class label), and CelebA~\cite{celeba} (gender as a class label). In CelebA, we used gender as a class label. Unlike person identity or other CelebA attributes, such as smiling and beard, several factors should be considered when selecting whether an observation is a female or a male. To demonstrate the effectiveness of our study, a classification task is required, in which several consistent factors are considered to classify an observation.

We predetermined the sizes of $c$ and $r$ for the three datasets. In MNIST and SVHN, $c$ $\in$ $\mathbb{R}^4$ and $r \in \mathbb{R}^8$. In CelebA, $c \in \mathbb{R}^8$ and $r \in \mathbb{R}^{16}$. Sec. S6 of the supplementary discusses the sizes of $c$ and $r$ in detail. In addition, Sec. S1 of the supplementary provides details of experiment information of our work.

\subsection{Informativeness}
\label{section:informativeness}
We measured the classification performance of our model and compared it with those of ResNet-18~\cite{resnet} and DeepCaps~\cite{deepcaps}. Our method shows similar accuracies on all three datasets, as shown in Table.~\ref{acc}. Our model can provide an explanation of the model's prediction without degradation in classification performance. We verified the informativeness of $c$ using quantitative and qualitative methods.

\begin{table}[t]
\centering
\caption{Classification accuracy of $p(y|r)$. We trained a classifier using $r$ of ours and the comparison methods. The classifier would have a random chance for test datasets if there is no class-relevant information in $r$ (lower is better).}
\begin{tabular}{@{}ccccc@{}}
\toprule
Architecture                                    &  & MNIST         & SVHN          & CelebA        \\ \midrule
Cycle-VAE~\cite{cyclevae} &  & 0.176         & 0.436         & 0.793         \\
ML-VAE~\cite{mlvae}       &  & 0.717         & 0.445         & 0.786         \\
LORD~\cite{lord}          &  & 0.099         & 0.163         & 0.517         \\
Ours                                            &  & 0.099         & 0.099         & 0.501         \\ \midrule
Random Chance                                   &  & 0.100         & 0.100         & 0.500         \\ \bottomrule
\end{tabular}
\label{acc_r}
\end{table}
\subsubsection{Quantitative Experiments}
By following the protocol from Cycle-VAE~\cite{cyclevae}, we trained a simple classifier to classify class labels from the residual representation $r$. This experiment was conducted to evaluate whether the class-relevant information is present in $r$. For comparison, we used recent class-supervised disentanglement methods~\cite{cyclevae,mlvae,lord}. The details of hyperparameters used for the comparison methods are provided in Sec. S9 of the supplementary.

\begin{figure*}[t]
\centering
\includegraphics[width=1\linewidth]{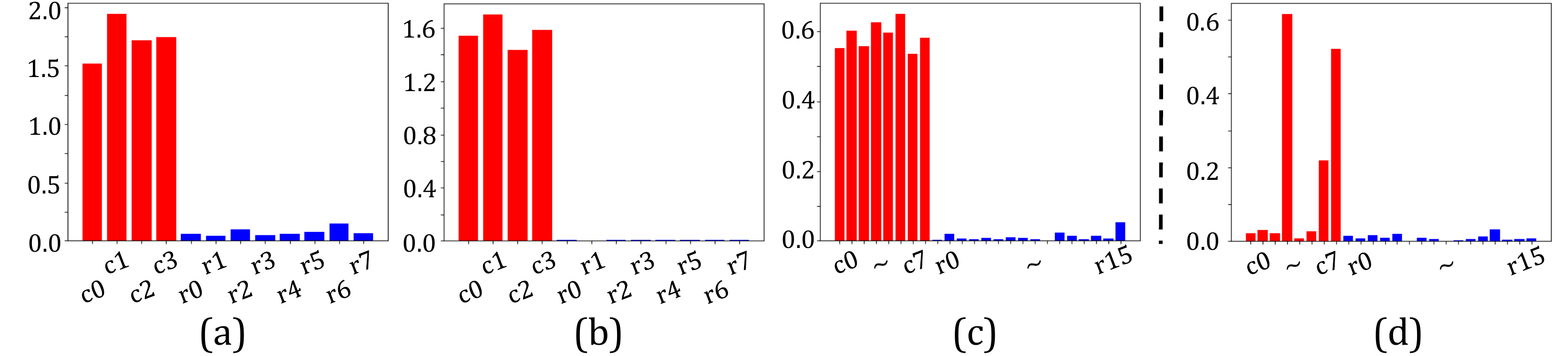}
\caption{Mutual information between $y$ and $c, r$ of (a) MNIST, (b) SVHN, (c) CelebA, and (d) CelebA w/o $\mathcal{L}_{\text{concept}}$ (higher is better for $c$; lower is better for $r$)}
\label{fig:ml}
\end{figure*}

As shown in Table.~\ref{acc_r}, only our method preserved a random chance in all the three datasets. This is because unlike the comparison methods, we did not agree with and implement the assumption that intra-class variation can be ignored. For datasets created for disentanglement learning, such as Cars3D~\cite{car3d}, SmallNorb~\cite{smallnorb}, KTH~\cite{kth}, etc., class-relevant intra-class variation is certainly small. However, for complex real datasets, class-relevant intra-class variation is typical (even large). Empirically, the classification accuracies of SVHN and CelebA in Table.~\ref{acc_r} show that the completely class-irrelevant $r$ cannot be created when assumed that intra-class variation can be ignored. For further analysis, we measured the mutual information between $y$ and $c$, $r$ of our model: $I(c;\: y)$ and $I(r;\: y)$. The $c0$ to $c7$ in Fig.~\ref{fig:ml} represent each element of the output vector of the class capsule. As shown in Fig.~\ref{fig:ml}(a, b, c), all the elements of $c$ of our method are strongly correlated to $y$, and all the elements of $r$ are uncorrelated to $y$ for the datasets. 

As an ablation study, we tested the importance of $\mathcal{L}_{\text{concept}}$. As shown in Fig~\ref{fig:ml}(d), some elements of $c$ obtained without $\mathcal{L}_{\text{concept}}$ show low correlations to $y$. $r$ still does not contain any class-relevant information, however, $\mathcal{L}_{\text{concept}}$ is required to encode class-relevant information to each element of $c$. Also, We tested the importance of $\mathcal{L}_M$ by replacing $C_C$ with ResNet-18. The result is discussed in Sec. S4 of the supplementary.

\begin{figure*}[t]
\centering
\includegraphics[width=0.8\linewidth]{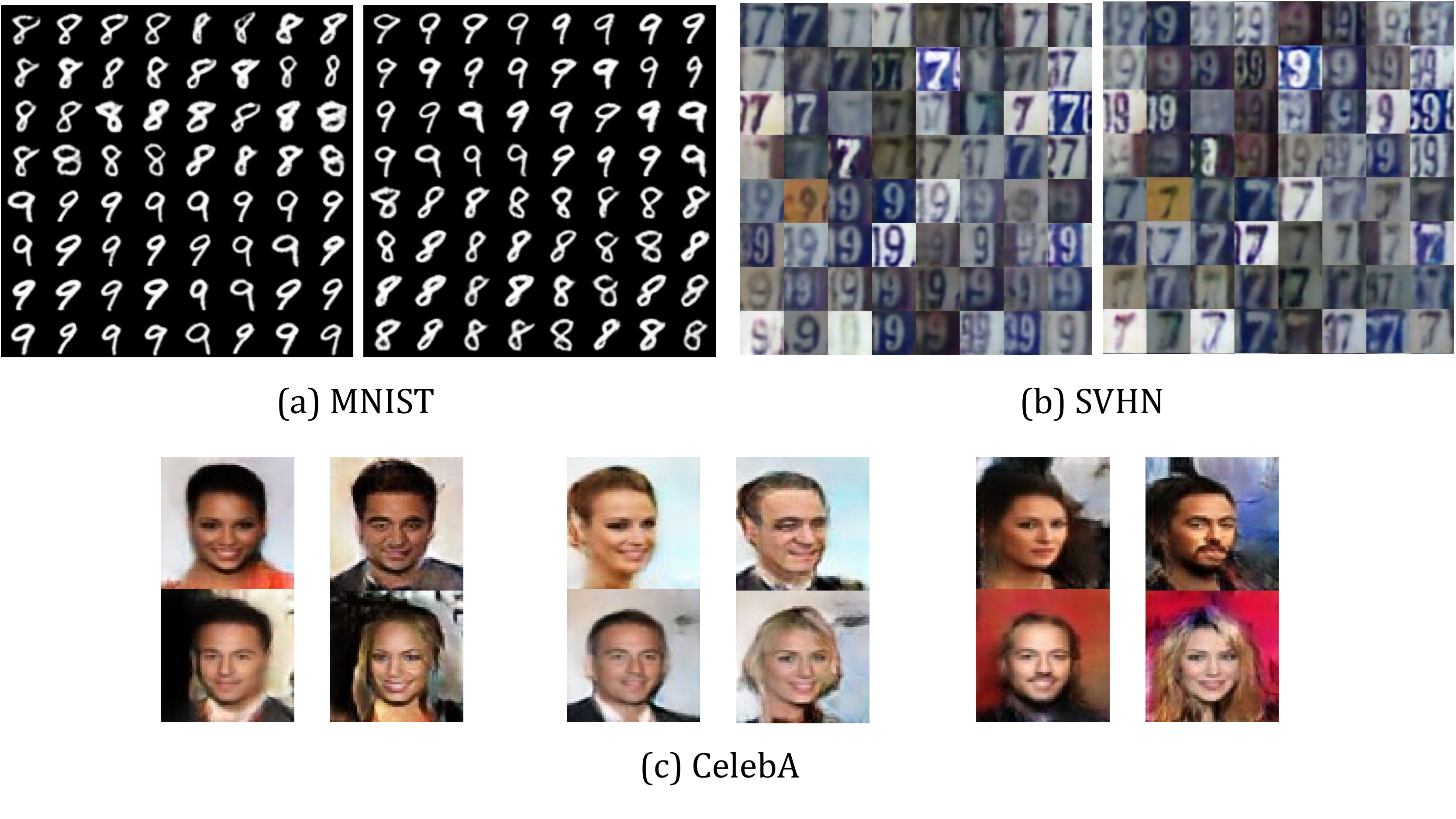}
\caption{Swapping. The images on the left of (a, b, c) are reconstructed using ($c^i$, $r^i$) and ($c^j$, $r^j$). The images on the right of (a, b, c) are generated by swapping $c$: ($c^j$, $r^i$) and ($c^i$, $r^j$).}
\label{fig:swapping}
\end{figure*}

\subsubsection{Qualitative Experiments} By swapping, we visually evaluated which factors of variation were encoded into $c$ and $r$ and whether they were semantically correct. From two test images of different class labels, we obtained ($c_i$, $r_i$) and ($c_j$, $r_j$), individually. Subsequently, we swapped $c_i$ and $c_j$ to generate new observations. The images on the left of Fig.~\ref{fig:swapping}(a, b, c) are generated using the original set: ($c_i$, $r_i$) and ($c_j$, $r_j$). The images on the right of Fig.~\ref{fig:swapping}(a, b, c) are generated by swapping: ($c_j$, $r_i$) and ($c_i$, $r_j$). It is clear that for MNIST, $r$ contains factors of variation such as thickness and skew. For SVHN, $r$ contains background color, font color, thickness, location, and skew. For CelebA, $r$ contains background color and a person's face feature. We believe that the results show semantically correct disentanglement. A detailed analysis of $c$ will be presented in Sec.~\ref{section:explanability}. In addition, the T-SNE~\cite{tsne} results of $r$ and $c$ of our method are shown in Sec. S7. The measured FID~\cite{fid} scores of the images by reconstruction, swapping, and random generation are provided in Sec. S8 of the supplementary.

\subsection{Distinctness}
\begin{figure*}[t]
\centering
\includegraphics[width=0.9\linewidth]{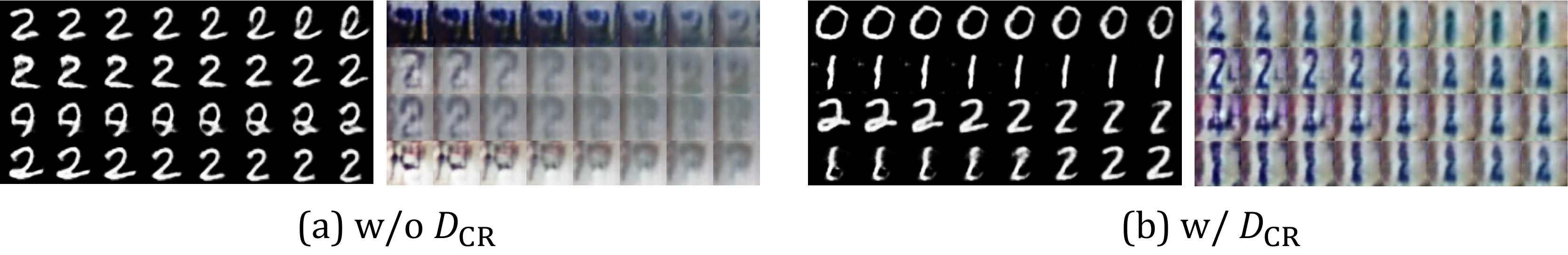}
\caption{Ablation study of $D_{\text{CR}}$. For (a), changes in rows overlap, whereas  changes in rows are distinct in (b).}
\label{fig:crd}
\end{figure*}

Distinctness means that each concept should be represented by a single variable. The variable type varies in each method, which can be a prototype or a latent feature~\cite{protopv2,senn}. In our method, the single concept is represented by a single element of the class capsule. To enforce it, we used an additional discriminator, $D_{\text{CR}}$. Fig.~\ref{fig:crd}(a) shows the images generated by $G$ trained without $D_{\text{CR}}$. Each row of image (a) indicates each element of the class capsule, and the eight images of the row are the results of linear interpolation between -1 and 1. In Fig.~\ref{fig:crd}(b), the change in each element is distinct from each other, whereas in Fig.~\ref{fig:crd}(a), the elements overlap. For example, the change in the left-half of the first row and the change in the right-half of the fourth row in MNIST of Fig.~\ref{fig:crd}(a) are the almost same. In such a case, the values for a single concept would be contradicting. This shows that $D_{\text{CR}}$ is required to enforce distinctness between the elements. The same trend was observed in CelebA. The qualitative and quantitative results of CelebA are given in Fig. S2 and Table. S3 of the supplementary. In addition, the importance of all the six components is discussed in Sec. S2 of the supplementary.

\subsection{Explainability}
\label{section:explanability}
\begin{figure*}[t]
\centering
\includegraphics[width=0.9\linewidth]{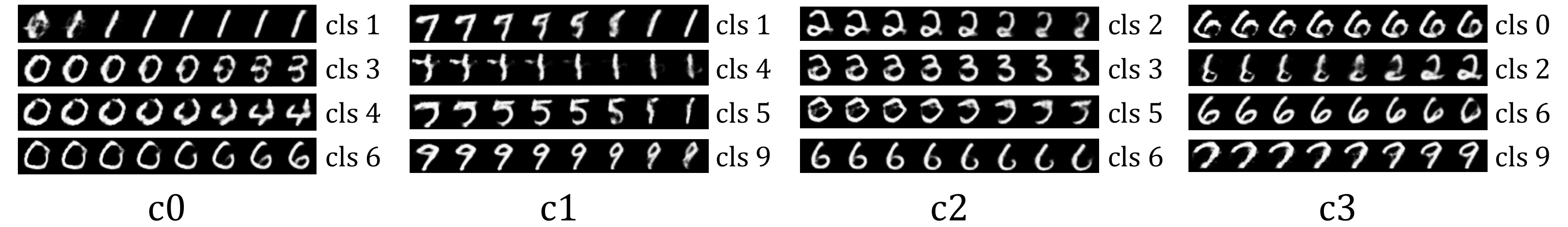}
\caption{Concepts learned for MNIST. c0: being a big circle, c1: being a straight line, c2: creating a small circle in the bottom part and c3: changing the upper part of the digit to a line.}
\label{fig:mnist_concept}
\end{figure*}

If explanations are provided based on concepts, these concepts should be human-understandable. In our setting, we demonstrate the learned concepts by linearly interpolating or analyzing data points of similar values. In Fig.~\ref{fig:mnist_concept}, the concept of each element is shown by linear interpolation. In MNIST, $r$ contains concepts such as thickness and skew, as described in Sec.~\ref{section:informativeness}. For $c$, the first element represents being a large circle. As the value approaches to $-1$, a majority of the digits are changed to zero. The second element represents being a straight line, and the third and fourth elements represent creating a small circle in the bottom part and changing the upper part of the digit to a line, respectively. By analyzing these results, we recognized that the factors relevant to MNIST classification are the size, number, and location of the circles and lines, and this finding also fits to SVHN in a very similar way (Sec. S5).

\begin{figure*}[t]
\centering
\includegraphics[width=0.9\linewidth]{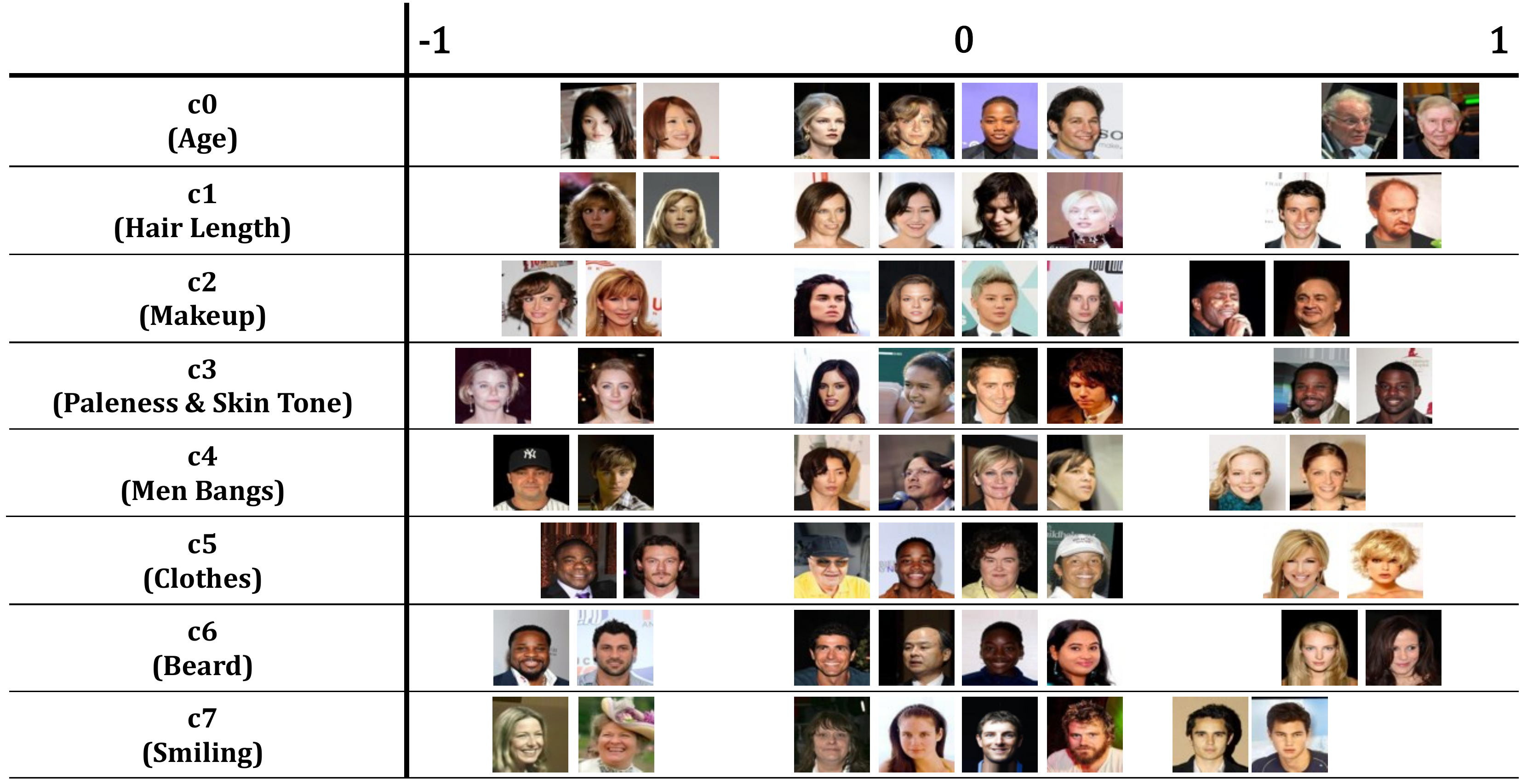}
\caption{Concepts learned for CelebA. For more details, $c2$ represents makeup. As the value approaches -1, people with heavy makeup appear. $c6$ represents a beard. As the value approaches -1, people who have a beard appear. Other concepts can be understood by verifying which image has a negative or positive value for each element.}
\label{fig:celeba_concept}
\end{figure*}

For CelebA, we show the concept of each element by analyzing a set of test images of certain values. We discovered the concepts such as age, hair length, makeup, paleness, skin tone, men hairstyle, clothes, beard, and smile, as shown in Fig.~\ref{fig:celeba_concept}. Case $c1$ represents hair length. As the value approaches 1, a person with extremely short hair appears. Case $c4$ represents men’s hairstyle. A person who has men bangs and perm typically exhibits a value less than 0. In case $c5$, the majority of men exhibit values less than 0. $c5$ is close to -1 when the person wears a suit and close to 1 when the person wears open-shoulder clothes.

We discovered an interesting phenomenon: In case $c0$, the element represents age. For persons appearing young, they typically have a value less than 0. Statistically, the number of females with values less than 0 was high, and the number of females with values greater than 0 was low. For males, the situation was vice versa. The model learned data bias from the CelebA dataset. When we analyzed the attribute named “young” of the CelebA dataset, we discovered an imbalance in the number of data, i.e., a ratio 2:1 (female:male). Similar to this case, we discovered an imbalance in CelebA attributes “pale skin” (3:1) and “smiling” (2:1), and these biases were encoded as class-relevant concepts ($c0$, $c3$, and $c7$). We discovered that our model can be used as a detector of hidden data bias; this will be investigated in future studies.

\begin{figure*}[t]
\centering
\includegraphics[width=0.9\linewidth]{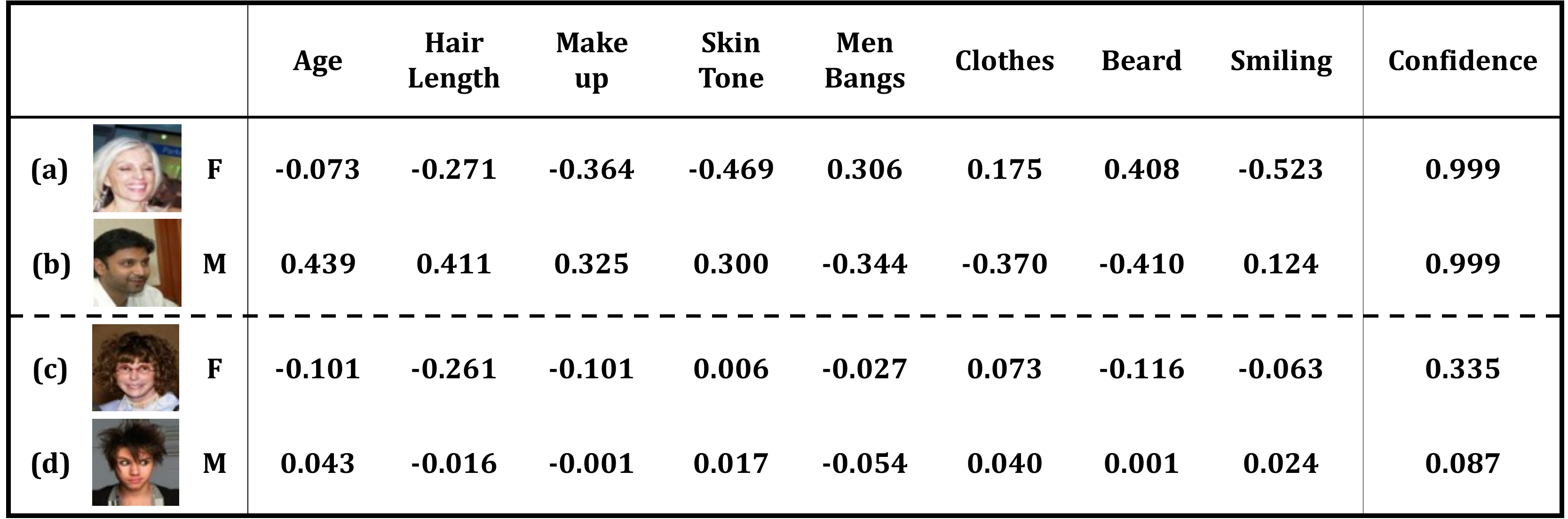}
\caption{Explanations generated by iCaps for four samples. In the cases of (a, b), iCaps is accurate in the predictions. On the contrary, in the cases of (c, d), iCaps misclassified with a low confidence score. By analyzing the values of each concept of the samples, we can understand the high and low confidence of iCaps in making the predictions.}
\label{fig:samples}
\end{figure*}

Herein, we demonstrate the explainability of our method using samples. In Fig.~\ref{fig:samples}, we present classification success cases of a female and a male, as well as misclassification cases of a female and a male. By analyzing the values of the concepts, we understood why the model classified Fig~\ref{fig:samples}(a) as a female with such high confidence. Fig~\ref{fig:samples}(a) was predicted as a female owing to observations of long hair, pale skin, no men bangs, no beard, and a smiling face. In the misclassification case (d), the model showed very low confidence in the classification because the model could not find a strong relevance to any concepts. 

\section{Conclusion and Future Work}
We propose a novel disentanglement method that the class-relevant subspace contains both class-relevant inter- and intra-class variation. Using the proposed method, we build a new interpretable model that provides explanations of the model's prediction based on class-relevant distinct concepts. 

In addition, the generator of our model can generate an image of the desired combination of the concepts. Hence, it can be used for data augmentation or additional explanations. Also, we will keep analyzing the possibility of our model as a detector of data bias. In future studies, we try to improve reconstruction ability and further, add a sentence generation phase at the end so that the model can generate an explanation as a sentence automatically.

\textbf{Acknowledgments.} This work was supported by the National Research Foundation of Korea (NRF) grant funded by the Korea government (Ministry of Science and ICT) [2018R1A2B3001628], the Brain Korea 21 Plus Project in 2020, Samsung Advanced Institute of Technology and Institute for Information $\&$ Communications Technology Planning $\&$ Evaluation (IITP) grant funded by the Korea government (MSIT) (No.2019-0-01367, BabyMind), and AIR Lab in Hyundai Motor Company through HMC-SNU AI Consortium Fund.

% egbib
\bibliographystyle{splncs04}
\bibliography{main}

\end{document}